\documentclass[]{spie}  %>>> use for US letter paper
\usepackage{amsmath,amsfonts,amssymb}
\usepackage{graphicx}
\usepackage{wrapfig}
\usepackage{cool}
\usepackage[colorlinks=true, allcolors=blue]{hyperref}

\title{Convolutional neural network based automatic plaque characterization from intracoronary optical coherence tomography images}

\author[1]{Shenghua He}
\author[2]{Jie Zheng}
\author[5]{Akiko Maehara}
\author[5]{Gary Mintz}
\author[6]{Dalin Tang}
\author[1,2,3,4]{Mark Anastasio}
\author[3]{Hua Li}
\affil[1]{Department of Computer Science and Engineering, Washington University in St. Louis, St. Louis, MO, USA}
\affil[2]{Department of Radiology, Washington University School of Medicine, St. Louis, MO, USA}
\affil[3]{Department of Radiation Oncology, Washington University School of Medicine, St. Louis, MO, USA}
\affil[4]{Department of Biomedical Engineering, Washington University in St. Louis, St. Louis, MO, USA}
\affil[5]{Cardiovascular Research Foundation, New York, NY}
\affil[6]{Worcester Polytechhic Institute, Worcester, MA}

\authorinfo{Further author information: (Send correspondence to Hua Li)\\Hua Li: E-mail: li.hua@wustl.edu, Telephone: 1 314 537 7145}
\begin{document} 
\maketitle

\begin{abstract}
Optical coherence tomography (OCT) can provide high-resolution cross-sectional images for analyzing superficial plaques in coronary arteries. Commonly, plaque characterization using intra-coronary OCT images is performed manually by expert observers. This manual analysis is time consuming and its accuracy heavily relies on the experience of human observers. Traditional machine learning based methods, such as the least squares support vector machine and random forest methods, have been recently employed to automatically characterize plaque regions in OCT images. Several processing steps, including feature extraction, informative feature selection, and final pixel classification, are commonly used in these traditional methods. Therefore, the final classification accuracy can be jeopardized by error or inaccuracy within each of these steps. In this study, we proposed a convolutional neural network (CNN) based method to automatically characterize plaques in OCT images. Unlike traditional methods, our method uses the image as a direct input and performs classification as a single-step process. The experiments on 269 OCT images showed that the average prediction accuracy of CNN-based method was 0.866, which indicated a great promise for clinical translation. 

\end{abstract}

% Include a list of keywords after the abstract 
\keywords{Convolutional neural network, automatic plaque characterization, optical coherence tomography}

\section{INTRODUCTION}
\label{sec:purpose}  % \label{} allows reference to this section
Optical coherence tomography (OCT) can achieve high-resolution and cross-sectional imaging of the internal microstructure in materials and biologic systems by measuring backscattered and backreflected light~\cite{fujimoto2000optical}. Commonly, for characterizing superficial plaques in inter-coronary arteries, the acquired OCT images are manually differentiated into four types: lipid tissue (LT), fibrous tissue (FT), mixed tissue (MT) and calcified tissue (CA)~\cite{prati2012expert}. However, this manual process is laborious and time consuming. The accuracy also heavily relies on the experience of human observers. To avoid these problems, methods for automatically characterizing plaque types in intracoronary OCT images should be developed.
%and can be unreliable when the observer is non-experienced.
%Thus, automatic plaque characterization in intracoronary OCT images is an urgent task in the clinical arena. 

Recently, traditional machine learning methodologies have been applied to automatically characterize plagues from intracoronary OCT images.~\cite{xiaoya2017, athanasiou2014}. For example, Xiaoya et al.~\cite{xiaoya2017} proposed a least square support vector machine (LS-SVM) based method to only classify LT and FT tissues for analyzing the plaque thickness. They first employed Otsu's thresholding based method~\cite{otsu1979threshold} to detect the whole plaque tissue area. Then, they selected the informative gray level co-occurrence matrices (GLCMs)~\cite{kekre2010image} and local binary patterns (LBPs) features~\cite{nanni2010local} for each tissue pixel, and inputed them into a LS-SVM-based classifier\cite{suykens2002least} for pixel classification. However, only 9 OCT images were processed in their experiment, resulting in possible overfitting, although the reported prediction accuracy was 0.896. Differently, Athanasiou et al.~\cite{athanasiou2014} employed a random forest (RF) based method~\cite{liaw2002classification} to classify plaque tissue into all four types: LT, FT, MT and CA. Their method consists of several steps, such as tissue area selection with Otsu's thresholding~\cite{otsu1979threshold} based method, pixel clustering with K-mean algorithm, informative feature selection based on wrapper feature selection (WRP)~\cite{hall2003benchmarking}, and pixel classification using a RF-based classifier. Although this method tries to characterize more tissue types in OCT images, the complex processing pipeline might prevent it from practical uses.

Currently, deep learning (DL) methods have had a profound impact on computer vision and image analysis applications, such as image classification~\cite{he2016deep,zhang2017crescendonet}, segmentation~\cite{badrinarayanan2015segnet}, image completion~\cite{yeh2016semantic} and so on. Convolutional neutral network (CNN) based deep neural nework, as the most commonly employed DL method, has the advantage of automatically and intensively extracting features directly from images. In this study, we employ a CNN-based DL method to automatically characterize plaque tissues from introcoronary OCT images and address the issues that limit traditional methods.

\section{Methods}
\label{sec:method}
\subsection{Overview of CNN-based method}
As shown in Figure~\ref{fig:method}, our CNN-based automatic plaque characterization method includes two steps: tissue area detection and CNN-based pixel classification. First, we used Otsu's automatic thresholding~\cite{otsu1979threshold} based method to detect the tissue area in an OCT image. Second, we used a CNN-based classifier to classify each pixel in the tissue area into five different tissue categories: LT, FT, MT, CA and background (BK). The BK pixel was defined as the pixel that did not belong to any of the other four tissue types. In the following subsections, we will explain these two steps in detail.
\begin{figure}[htbp]
\vspace{0in}
\centering
\includegraphics[width=0.80\textwidth,height=0.28\textheight]{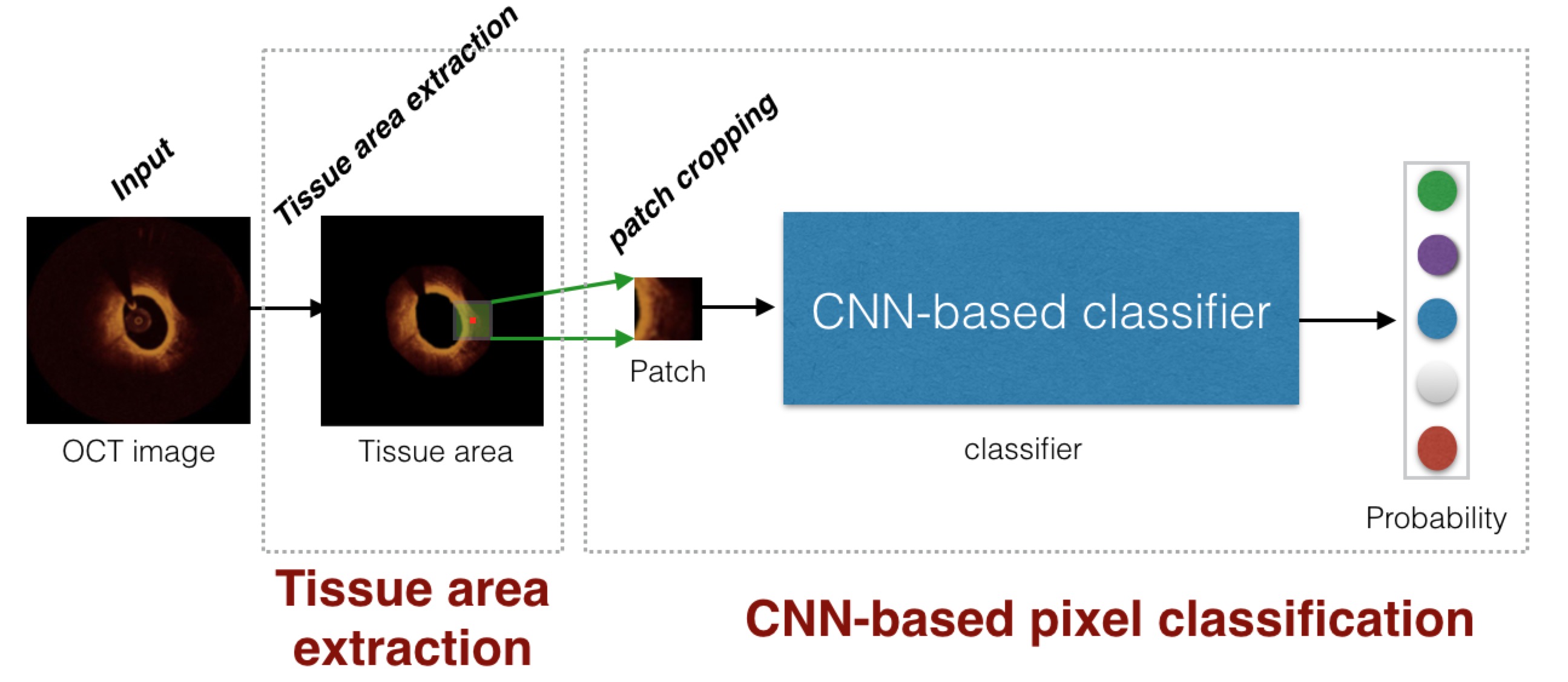}
\vspace{-0.05in}
\begin{center}
\caption{Overview of the CNN-based method}\label{fig:method}
\label{fig:method}
\end{center}
\vspace{-0.2in}
\end{figure}

\subsection{Step 1: Tissue area extraction}
The individual A-lines acquired by an OCT systems contain the information of the reflected optical energy as a function of time~\cite{fercher2010optical}. These A-lines are stored sequentially in a 2-D polar OCT image $I$ with each element corresponding to a polar intensity data. In a polar OCT image $I$, the top part corresponds to the area near gravitational center of the tissue, while the bottom part corresponds to the outer area outside the tissue. In each of these images, there are some catheter artifact pixels located outside the tissue area. In order to reduce their interferences on the accuracy of the pixel classification~\cite{athanasiou2014}, we first need to remove these catheter pixels and keep only the tissue area. This tissue area extraction procedure includes two steps: lumen border detection and border expansion.

%The OCT images used in our study are Cartesian-coordinate OCT images. For an image $I(x,y)$, where $(x,y)$ represent a 
%coordinate in Cartesian coordinates, 
%To effectively detect the lumen border in these OCT images, we first used linear transformation to transform $I(x,y)$ to 

To detect the lumen border, we first performed Otsu's automatic thresholding~\cite{otsu1979threshold} to remove catheter artifact pixels. With such procedure, we obtained a binary image that contains only the zero pixels and nonzero pixels. Afterwards, we scanned each column in $I$ from the top (gravitational center) to the bottom (outer area), and stored the first nonzero pixel in each of these columns. Finally, these stored nonzero pixels were connected to form the detected lumen border.

After extracting the lumen border (inner border), we expanded $1.5$ mm, as presented in the reference~\cite{athanasiou2014}, starting from this border towards the bottom (outer area), and obtained another border (outer border). The area between these two borders in the original polar OCT image was considered as the detected tissue area. Finally, in order to apply CNN-based classifier to these polar OCT images for pixel classification, we transformed these images from polar to Cartesian coordinates. Due to the border extension, some background pixels were included in this tissue area. As a result, we classified all pixels in this OCT image into 5 tissue types: LT, FT, MT, CA, and BK.

\subsection{Step 2: CNN-based pixel classification}
Having the extracted tissue area, we next employed a CNN-based method to classify each pixel in this tissue area into one of the five tissue types: LT, FT, MT, CA and BK. As shown in Figure \ref{fig:method}, the input of the classifier is an image patch with the to-be-classified pixel at the center of this patch, and the classifier's outputs are five scores which denote the probabilities that each to-be-classified pixel belongs to the LT, FT, MT, CA, and BK classes, respectively. 

Our CNN-based classifier can be modeled as a nonlinear function, $\boldsymbol{p_y} = P(\boldsymbol{x};\boldsymbol{\theta})$, which maps a 2-D image patch $\boldsymbol{x} \in \mathbb{R}^{H\times H}$ to a vector  $\boldsymbol{p_y} = (p_{1}, p_{2}, ..., p_{5})^T$, where $H \times H$ is the size of the OCT image patch. Here, each $p_i \in [0,1] (i = 1,2,3,4,5)$ denotes the probability of the current image patch $\boldsymbol{x}$ belonging to the $i$-th tissue category. The mapping also depends on the set of parameters $\boldsymbol{\theta} = \{\theta_{1},\theta_{2}, ...,\theta_{K}\}$, where $K$ was the total number of trainable parameters in our classifier.

The network architecture design, network training strategy and data preprocessing strategy of our CNN-based classifier are presented in Sections~\ref{sect:architecture},~\ref{training},~\ref{sec:gener}, respectively. The network architecture design (in Section \ref{sect:architecture}) determines the classifier mapping model $P(\boldsymbol{x};\boldsymbol{\theta})$ and specifies $K$. The network training strategy (in Section \ref{training}) describes how to configure values for all the parameters in $\boldsymbol{\theta}$. The data preprocessing strategy (in Section \ref{sec:gener}) introduces the way we generated the training sample $(\boldsymbol{x}, \boldsymbol{p_y})$ for our classifier training, and validation sample for classifier validation.

\subsubsection{An architecture of CNN-based classifier}\label{sect:architecture}
\begin{figure*}[htbp]
\centering
\vspace{-0in}
\includegraphics[width=0.90\textwidth,height=0.25\textheight]{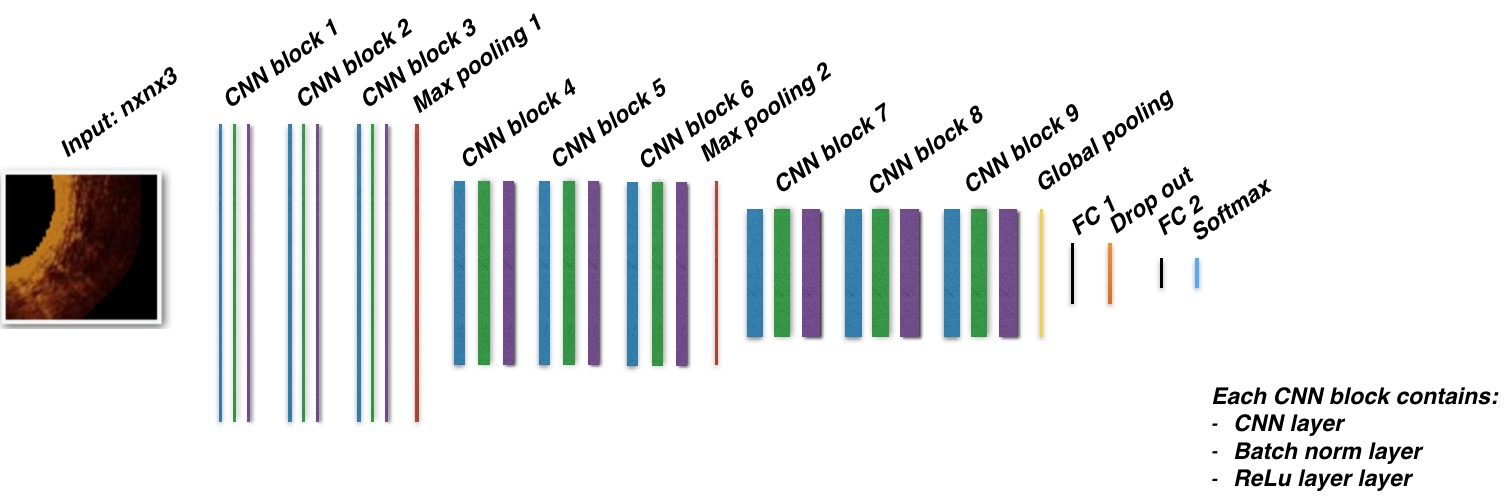}
\vspace{-0.0in}
\begin{center}
\caption{The architecture of our proposed CNN-base classifier}
\label{fig:architecture}
\end{center}
\vspace{-0.1in}
\end{figure*}

Generally, a CNN-based deep neural network consists of a number of convolutional (CONV) layers followed by a number of fully connected (FC) layers. The CONV layers extract the high-level features from an image patch, then the classification is performed on these features by use of the FC layers. In this study, trial and error method was used to identify the CNN architecture to avoid the overfitting problem, and the number of CONV layers and that of FC layers were set to 9 and 2, respectively. 

The network architecture employed in this study is shown in Figure~\ref{fig:architecture}. For description convenience, we defined a CONV block as a sequence of layers, which consisted of a CONV layer, a batch normalization layer, and a ReLu layer. As shown in \ref{fig:architecture}, our network architecture contained 9 CONV blocks and 2 FC layers. 2 max pooling layers were placed after the 3rd and the 6th CONV blocks, respectively. 1 global pooling layer was placed after the 9th CONV block.  The spatial support of the filter in each of the CONV layers was set as $3\times3$ pixels. The number of the filters in first three CONV layers was set to $32$. In order to compensate for the information loss caused by max pooling, the number of filters in 2nd three CONV layers and 3rd three CONV layers were set to $64$ and $128$, respectively. Two FC layers followed the global pooling layer. The first FC layer included 512 neurons and the second one included 5 neurons. One dropout was set between these two FC layers with a dropout ratio 0.5 to further avoid overfitting. A softmax layer was placed at the end of our classifier to produce probability scores. The input of our CNN-based clssifier was OCT image patch (described in \ref{sec:gener}). The outputs of the classifier were 5 probability-like scores.

For this network architecture, the CNN mapping function $P(\boldsymbol{x};\boldsymbol{\theta})$ is then fixed with $K=550725$.

\subsubsection{The training strategy of our CNN-based classifier}
\label{training}

Given a set of training data, the goal of classifier training is to find a set of parameters $\boldsymbol{\theta}$ that minimizes a loss function that quantifies the average error between the true category of the training data and the category predicted by the classifier. 

In this study, the training dataset consisted of $N$ image patches $\{\boldsymbol{x}^{(i)}\}, i = 1, 2, ..., N$. Each image patch $\boldsymbol{x}^{(i)}$ was categorized as one of the five tissue types: BK, LP, FT, MT and CA, and corresponds to a one-hot label vector $\boldsymbol{y}^{(i)}$ as defined in Table \ref{tab:class}. The cross-entropy loss function $L_{\{\boldsymbol{x}^{(1)},...,\boldsymbol{x}^{(N)}\}}(\boldsymbol{\theta})$ was employed: 

\begin{align} 
\label{eq1}
   L_{\{\boldsymbol{x}^{(1)},...,\boldsymbol{x}^{(N)}\}}(\boldsymbol{\theta}) = -\frac{1}{N}\sum_{i=1}^{N} w_i \boldsymbol{y}^{(i)T} \log P(\boldsymbol{x}^{(i)};\boldsymbol{\theta}),
\end{align}
 where $w_i$ is the weight for the $i$-th training data. For a given $\boldsymbol{x}^{(i)}$, if this patch belongs to class $j$, $w_i$ will be defined in Eq.~\ref{eq2}, 
\begin{align} 
\label{eq2}
w_i = \frac{\frac{1}{M_j}}{\sum_{j=0}^{4}\frac{1}{M_j}},
\end{align}
where $M_j$ is the number of training data that belong to class $j (j =1,2,...,5)$. The weight $w_i$ was utilized to compensate for the fact that the training data with minor classes have less opportunities to update the classifier parameters.

The training of our classifier can be defined as an nonlinear optimization problem:
\begin{align} 
\label{eq3}
   \hat{\boldsymbol{\theta}} = \underset{\boldsymbol{\theta}}{\arg\min}L_{\{\boldsymbol{x}^{(1)},...,\boldsymbol{x}^{(N)}\}}(\mathbf{\boldsymbol{\theta}}).
\end{align}
We employed the momentum stochastic gradient descent (SGD)~\cite{bottou2010large}optimizer to solve the Eq.~\ref{eq3}. The parameter $\theta_i$ was updated as:
\begin{align} 
\label{eq4}
   \theta_{i}^{t+1} = \lambda \theta_{i}^{t} + (1-\lambda)  (-\eta \pderiv{L^t}{\theta_{i}}),
%= W_{i}^{t}+\pderiv{E^t}{W_{i}}
\end{align}
where $\theta_{i}^{t}$ denotes the value of $\theta_{i}$  at $t$-th iteration,  $\eta$ is the learning rate which controls the speed of update, and momentum $\lambda \in (0,1]$ determines the degree that the previous gradients are incorporated into the current update. $\pderiv{L^t}{\theta_{i}}$ is the gradient provided by one batch of training data at the $t$-th iteration, which can be calculated by use of the backpropogation algorithm~\cite{hecht1988theory}. In this study, the learning rate $\eta$ and momentum $\lambda$ were set to 0.0001 and 0.9, respectively. The batch size was set to 216.

\begin{table}[ht]
\caption{One-hot true label vector for the five tissue types} 
\label{tab:class}
\begin{center}       
\begin{tabular}{|l|l|} %% this creates two columns
%% |l|l| to left justify each column entry
%% |c|c| to center each column entry
%% use of \rule[]{}{} below opens up each row
\hline
\rule[-1ex]{0pt}{3.5ex}  Class (Tissue type) & Label $\mathbf{y^T}$ \\
\hline
\rule[-1ex]{0pt}{3.5ex}  class 1 (BK) & $(1,0,0,0,0)$   \\
\hline
\rule[-1ex]{0pt}{3.5ex}  class 2 (LP) & $(0,1,0,0,0)$  \\
\hline
\rule[-1ex]{0pt}{3.5ex} class 3 (FT) & $(0,0,1,0,0)$   \\
\hline
\rule[-1ex]{0pt}{3.5ex}  class 4 (MT) & $(0,0,0,1,0)$    \\
\hline
\rule[-1ex]{0pt}{3.5ex}  class 5 (CA) & $(0,0,0,0,1)$  \\
\hline
\end{tabular}
\end{center}
\end{table} 

\subsubsection{Data preprocessing}
\label{sec:gener}

The training and validation data employed in our classifier training were image patches generated from Cartesian-coordinate OCT images. 

At each iteration of parameter update defined in Eq.~\ref{eq4}, we randomly extracted a patch with size $51 \times 51$ from each of OCT images in the training set. Each image patch and its corresponding class label were paired as a training sample. These generated training samples were formed as a training batch to update the parameters. To mitigate overfitting, we augmented the OCT images for every 200 iterations by using image rotation with a random degree in range $[0,50]$. 

Additionally, at each iteration, a set of validation samples were generated by randomly extracting $1000$ $51 \times 51$ image patches from each of the OCT images in the validation set. These validation samples were used for model selection during the classifier training. The training of our CNN-based classifier took about 3 millions of iterations. During this period, the parameters that resulted in the best prediction accuracy on the set of validation samples were considered as best parameters, and these parameters will be used for the performance evaluation of the classifier.

\section{Experimental results}
\label{sec:result}

\begin{figure}[!htb]
	 \vspace{-0.0in}
        \centering
        \includegraphics[width=0.7\linewidth, height=0.06\textheight]{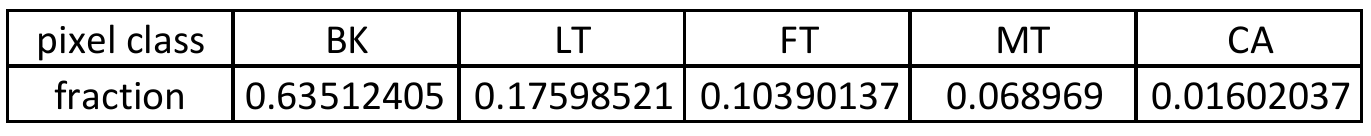}
        \caption{The fractions of pixels in each class}
        \label{fig:dis}
	 \vspace{-0.10in}
\end{figure}

The image set used in our experiment contained 269 OCT images acquired from 22 patients. Each OCT image had a ground truth counterpart, which indicated the class label for every pixel in this OCT image. These ground truth data were manually established by expert observers. The fractions of pixels in each class in the whole ground truth data are shown in Figure~\ref{fig:dis}. 

The training and validation of our CNN-based classifier were performed on a NVIDIA Titan X GPU with 12GB of VRAM. Software packages used in our experiments included Python 3.4, Keras 2.0 and tensorflow 1.0. In order to evaluate our CNN-based method, we first randomly shuffled OCT images and evenly divided them into 5 non-overlap subsets. Then we performed the 5-fold cross validation~\cite{refaeilzadeh2009cross} method on these image subsets to avoid the evaluation variance.
%During the training period, the model with best average validation prediction accuracy will be stored as the best model. %Afterwards, each 

\begin{figure}[!htb]
	 \vspace{-0.0in}
        \centering
        \includegraphics[width=0.7\linewidth, height=0.06\textheight]{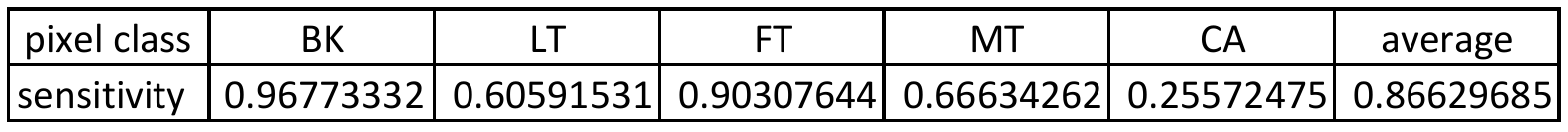}
        \caption{Sensitivity performance}
        \label{fig:acc}
	 \vspace{0.00in}
\end{figure}

In this study, we used a sensitivity metric to evaluate the classification of each tissue type, which was defined by:
\begin{align}
\label{eq5}
  Sen=\frac{TP}{TP+FN}
\end{align}
where $TP$ is the number of true positive predictions, while $FN$ is the false negative predictions. As shown in Figure ~\ref{fig:acc}, the average prediction sensitivities for the background and FT tissue classes can both achieve over 0.9. For tissue LP and MT, the average prediction sensitivities are over 0.6. However, the prediction sensitivity for CA tissue type is lowest, this might be due to the tiny ratio ($0.016$ shown in Figure \ref{fig:dis}) of the CA pixels in the dataset.

\begin{figure}[!htb]
%	 \vspace{-0.35in}
        \centering
        \includegraphics[width=0.60\linewidth, height=0.30\textheight]{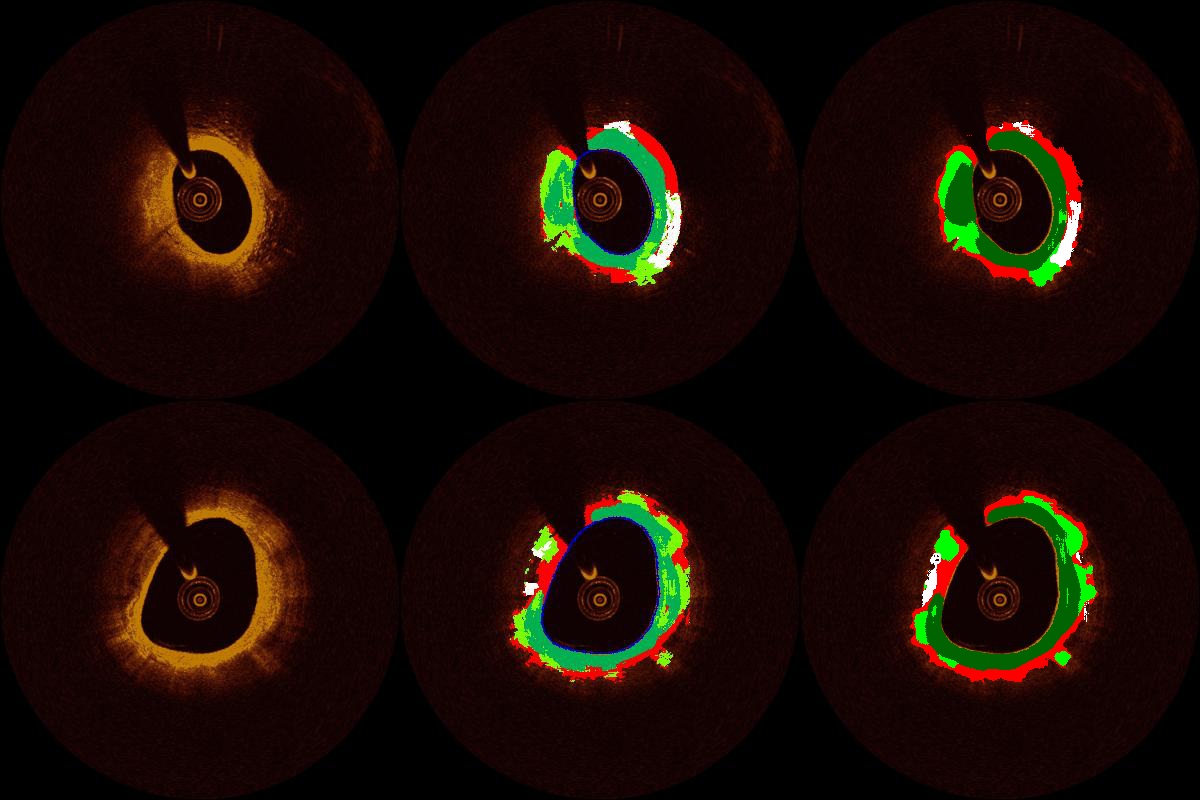}
        \caption{The tissue classification on two example OCT images. FT: dark green; LT: red; MT: light green; CA: white. Each column shows respectively the original OCT images, ground truth characterization images and the CNN-based characterization results in our experiments, respectively.}
        \label{fig:visual_segment}
%        \vspace{-0.3in}
\end{figure}

Figure \ref{fig:visual_segment} gives two classification examples. It shows that the characterization results with our proposed method are close to the ground truth ones.

\section{Conclusion}
\label{sec:conclusion}
In this study, we developed a CNN-based method for automatic plaque characterization on OCT images. Our method can extract informative features directly from OCT image patches for pixel classification. Experimental results showed that the average pixel prediction accuracy was 0.866. We also demonstrated that our proposed method can detect the background and FT tissue regions with a sensitivity of over 0.9. These results show that the CNN-based automatic tissue segmentation method holds great promise for clinical translation. In future, we will acquire more OCT images and use them to retrain our CNN classifier, in order to improve the classification accuracy for LT, MT and CA classes.
\bibliography{SPIE_manuscript_7.1} % bibliography data in report.bib
\bibliographystyle{spiebib} % makes bibtex use spiebib.bst

\end{document}